\documentclass[11pt,a4paper]{article}
\pdfoutput=1
\usepackage{url}
\usepackage[hyperref]{eacl2021}
\usepackage{times}
\usepackage{latexsym}
\usepackage{amsmath}

\usepackage{float}

\usepackage{microtype}
\usepackage{graphicx}
\usepackage{subfigure}

\aclfinalcopy 

\makeatletter
\newcommand{\printfnsymbol}[1]{%
  \textsuperscript{\@fnsymbol{#1}}%
}
\makeatother

\title{Team Phoenix at WASSA 2021: Emotion Analysis on News Stories with Pre-Trained Language Models}

\author{Yash Butala\thanks{* Equal Contribution} , Kanishk Singh\printfnsymbol{1}, Adarsh Kumar\printfnsymbol{1} and Shrey Shrivastava\\
\\
  Indian Institute of Technology Kharagpur, India\\
  \texttt{\{yashbutala,kanishksingh,adarshkumar712\}@iitkgp.ac.in} \\
  \texttt{shrivastava.shrey@iitkgp.ac.in}
  }

\date{}

\begin{document}
\maketitle
\begin{abstract}
Emotion is fundamental to humanity. The ability to perceive, understand and respond to social interactions in a human-like manner is one of the most desired capabilities in artificial agents, particularly in social-media bots. Over the past few years, computational understanding and detection of emotional aspects in language have been vital in advancing human-computer interaction.  The WASSA Shared Task 2021 released a dataset of news-stories across two tracks, Track-1 for Empathy and Distress Prediction and Track-2 for Multi-Dimension Emotion prediction at the essay-level. We describe our system entry for the WASSA 2021 Shared Task (for both Track-1 and Track-2), where we leveraged the information from Pre-trained language models for Track specific Tasks. Our proposed models achieved an Average Pearson Score of \textbf{0.417}, and a Macro-F1 Score of \textbf{0.502} in Track 1 and Track 2, respectively. In the Shared Task leaderboard, we secured \textbf{4\textsuperscript{th}} rank in Track 1 and \textbf{2\textsuperscript{nd}} rank in Track 2.
\end{abstract}

\section{Introduction}
Sentiment analysis over texts has been a widely researched area in NLP. The number of papers published in sentiment analysis related domains has increased from 37 papers in 2000 to 6996 in 2016 \cite{DBLP:journals/corr/MantylaGK16}. Sentiment analysis is a trending research topic, possibly due to its applications that automatically collect and analyze a large corpus of opinions with text mining tools. From the conventional task of predicting polarity as positive, negative or neutral, the researchers are now increasingly focused on sophisticated tasks such as emotion recognition, aspect level sentiment analysis, intensity prediction, etc. 

Recently, the researchers started exploring more sophisticated models of human emotion on a larger scale. Several datasets and corpora have been curated in this domain, such as   \cite{SemEval2018Task1}, \cite{tales_data}, and larger datasets like \cite{demszky-etal-2020-goemotions}.
\cite{buechel-etal-2018-modeling} presented an interesting computational work distinguishing between multiple forms of empathy, empathic concern, and personal distress. This data of empathic concern and personal distress along with Multi-dimension Emotions Labelling on news-stories across seven classes, namely: sadness, fear, neutral, anger, disgust, joy, surprise, has been released a Shared Task \cite{tafreshi-etal-2021} in WASSA-2021 Workshop as Two Tracks\footnote{refer Section \ref{Task} for more details}.


In this paper, we describe our system entry for both the tracks of WASSA 2021 Shared Task.
The primary contributions of the paper are as follows:
\vspace{-0.25cm}
\subsection*{Track 1:}
\begin{itemize}
	\item We demonstrate the efficacy of multi-tasking through parameter sharing which further strengthens the belief that empathic concern and personal distress are co-related.
	\vspace{-0.25cm}
	\item We amalgamate the information from sentence embeddings with normalized additional information to finally predict the empathic concern and personal distress using regression. 
\end{itemize}
\vspace{-0.50cm}
\subsection*{Track 2:}
\begin{itemize}
    \item We provide a comparative analysis of generation modelling against classification modelling for the task of Emotion Prediction. 
	\vspace{-0.25cm}
	\item We illustrate the efficiency of Task Specific Incremental Fine-Tuning approach \ref{EMO} on Pre-Trained Models for a small sized dataset.
\end{itemize}

\section{Related Works}
Pre-trained language models have proved to be a breakthrough in analyzing a person's emotional state. We now describe briefly some of these highly influential works.

\subsection{Pre-trained Language Models}

Over the past few years, pre-trained language models have progressed greatly in learning contextualized representations. Transformer \cite{Vaswani}, first proposed for machine translation, has enabled faster learning of complex representations of text. GPT \cite{GPT2}, BERT \cite{DBLP:journals/corr/abs-1810-04805}, RoBERTa \cite{Roberta}, XLNet \cite{xlnet} all leverage transformer architecture along with statistical tokenizers.  ELECTRA \cite{clark2020electra} a recent generator-discriminator-based pre-training approach offers a competitive performance despite requiring lesser compute. Domain specific language models also leads to a significant performance gain \cite{vaidhya-kaushal-2020-iitkgp, biobert, scibert}. 

\subsection{Emotion Recognition}
Emotion recognition through facial expressions and speech data has been the subject of extensive study in the past.
\cite{face-emo1} presents an approach for recognition of seven emotional states based on facial expressions. \cite{audio-emo1} utilizes a novel deep dual recurrent encoder model to obtain a better understanding of speech data using text data and audio signals simultaneously.

For text, various approaches have been proposed for emotion recognition.
\citet{Emotion-Classification:SVM} proposed an SVM-based approach for predicting opinions on news headlines.
\citet{bert-emo} paper analyses the efficacy of utilizing transformer encoders for detecting emotions.
\citet{Practical} demonstrates the practical efficiency of large pre-trained language models for Multi-Emotion sentiment classification.

\subsection{Computation of Empathy}

Empathy and distress are core components of a person's emotional state, and there has been a growing interest in computational approaches to model them. 
Considering language variations across different regions, empathy and distress can also vary with demographics \cite{demography-1-vary, demography-2-vary}, and recently \cite{empathy-bert-eacl-2021} proposed a demographic-aware empathy modelling framework using BERT and demographics features.

Understanding empathy and distress are crucial for analyzing mental health and providing aid. 
Recently \citep{empathy-mental-health-understanding-2020} explored language models for identifying empathetic conversations in the mental health support system. 

\section{Task and Dataset Description}
\label{Task}
Our experiments' data consists of the emotion-labels to news stories released as part of the WASSA 2021 shared task. The dataset provided \cite{buechel-etal-2018-modeling} contained essays of 300-800 characters length, Batson empathetic concern, and personal distress scores along with other additional demographic and personality information.

The training corpus of WASSA-2021 shared task consists of 1860 training pairs containing seven emotion labels, namely: sadness, fear, neutral, anger, disgust, joy, surprise. The dataset also includes person-level demographic information (age, gender, ethnicity, income, education level) and personality information. We normalized these information before using in our model for Track 1. We excluded this information in Track 2 model. 

\begin{table}[h]
\centering
\begin{tabular}{|p{2.0cm}|p{1.5cm}|p{1.5cm}|}
 \hline
 \textbf{Emotion} & \textbf{Train} & \textbf{Dev}\\
 \hline
  sadness &  647 & 96\\
  anger & 349  & 76\\
  neutral & 275 & 31\\
  fear & 194 & 25\\
  surprise & 164 & 14\\
  disgust & 149 & 14\\
  joy & 82 & 12\\
  \hline
  \textbf{Total} & \textbf{1860} & \textbf{270}\\
  \hline
\end{tabular}
\caption{Composition of Training and Development dataset}
\label{Dataset}
\end{table}

The objective of the Track-1 is to predict the Batson empathic concern and personal distress using the essay and any of the additional information to improve Pearson corelation between the predicted labels and gold standard labels. The task can formally be described as following:


\nocite{kaushal-vaidhya-2020-winners, vaidhya-kaushal-2020-iitkgp}
\paragraph{Empathic concern and personal distress prediction:} Given a paragraph $t$, additional information $i$,  learn a model:\\ $g(t,I)\rightarrow (x,y)$ where $x \in R\textsuperscript{+}$ and $y \in R\textsuperscript{+}$.
\\

\noindent The Track 2 is formulated as essay-level  Multi-Dimension emotion prediction task. It is defined formally as:  
\paragraph{Emotion Prediction:} Given a paragraph $t$, the classification task aims to learn a model $g(t)\rightarrow \{l_{1},l_{2},..,l_{k}\}$ where $l_{i}$ is a label $\in \{sadness,anger,..etc\}$.
\\

\section{Approach}
\subsection{Empathy and Distress Prediction Model}
\label{EMP}
\begin{figure}[htp]
    \centering
    \includegraphics[width=0.82\linewidth]{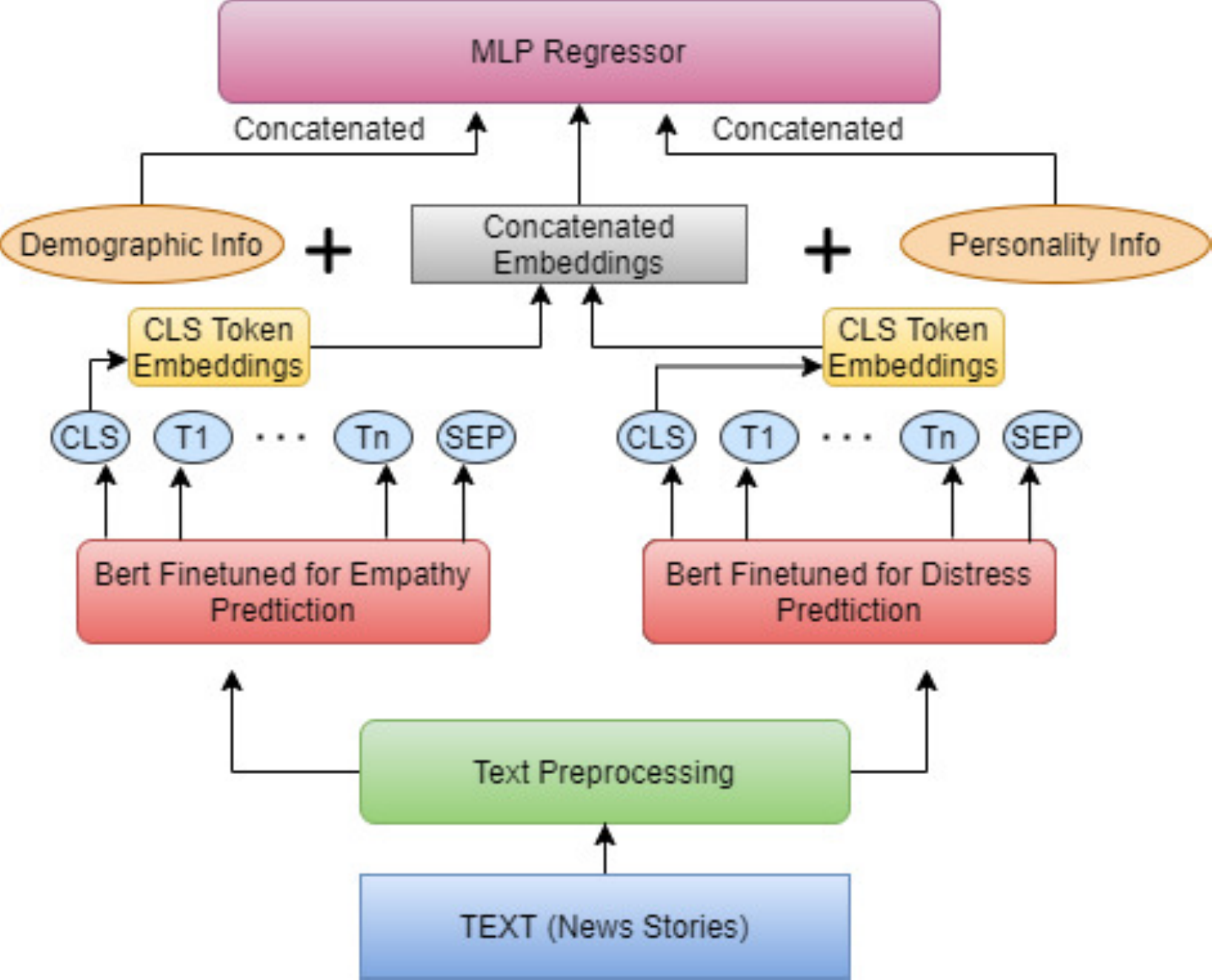}
    \caption{}
    \label{fig:a}
\end{figure}

 The system architecture for Empathy and Distress Prediction is shown in Figure \ref{fig:a}. The approach is primarily based on fine-tuning pre-trained language models for down-stream tasks. We enforce the technique of hard-parameter sharing through concatenation of BERT-fine-tuned embeddings trained separately for Empathy and Distress Prediction. The final shared parameters are then concatenated with the scaled demographic and personality features given in the dataset. These separately fine-tuned BERT-embeddings for distress and empathy prediction are then concatenated with rest of the features before feeding them to the regression models. This parameter shared multi-task framework \cite{kaushal-vaidhya-2020-winners} allows for the use of the same model, loss function, and hyper-parameters for the task of empathy prediction as well as Distress Prediction. 

{\small
\[
    MSE(y_{true}, y_{pred}) = \sqrt{(\frac{1}{n})\sum_{i=1}^{n}(y_{true_{i}} - y_{pred_{i}})^{2}}
\]

\begin{equation*}
  r = 
  \frac{ \sum_{i=1}^{n}(y_{true_i}-\bar{y_{true}})(y_{pred_i}-\bar{y_{pred}}) }{
        \sqrt{\sum_{i=1}^{n}(y_{true_i}-\bar{y_{true}})^2}\sqrt{\sum_{i=1}^{n}(y_{pred_i}-\bar{y_{pred}})^2}}
\end{equation*}
}

where $y_{true}$ are the gold-standard and $y_{pred}$ being the predicted scores for empathy and distress. The final Pearson correlation score used for final evaluation was: 
\[
    r_{avg} = \frac{r_{empathy}+r_{distress}}{2}
\]

\subsection{Emotion Label Generation Model}

\label{EMO}
\begin{figure}[H]
    \centering
    \includegraphics[width=0.7\linewidth]{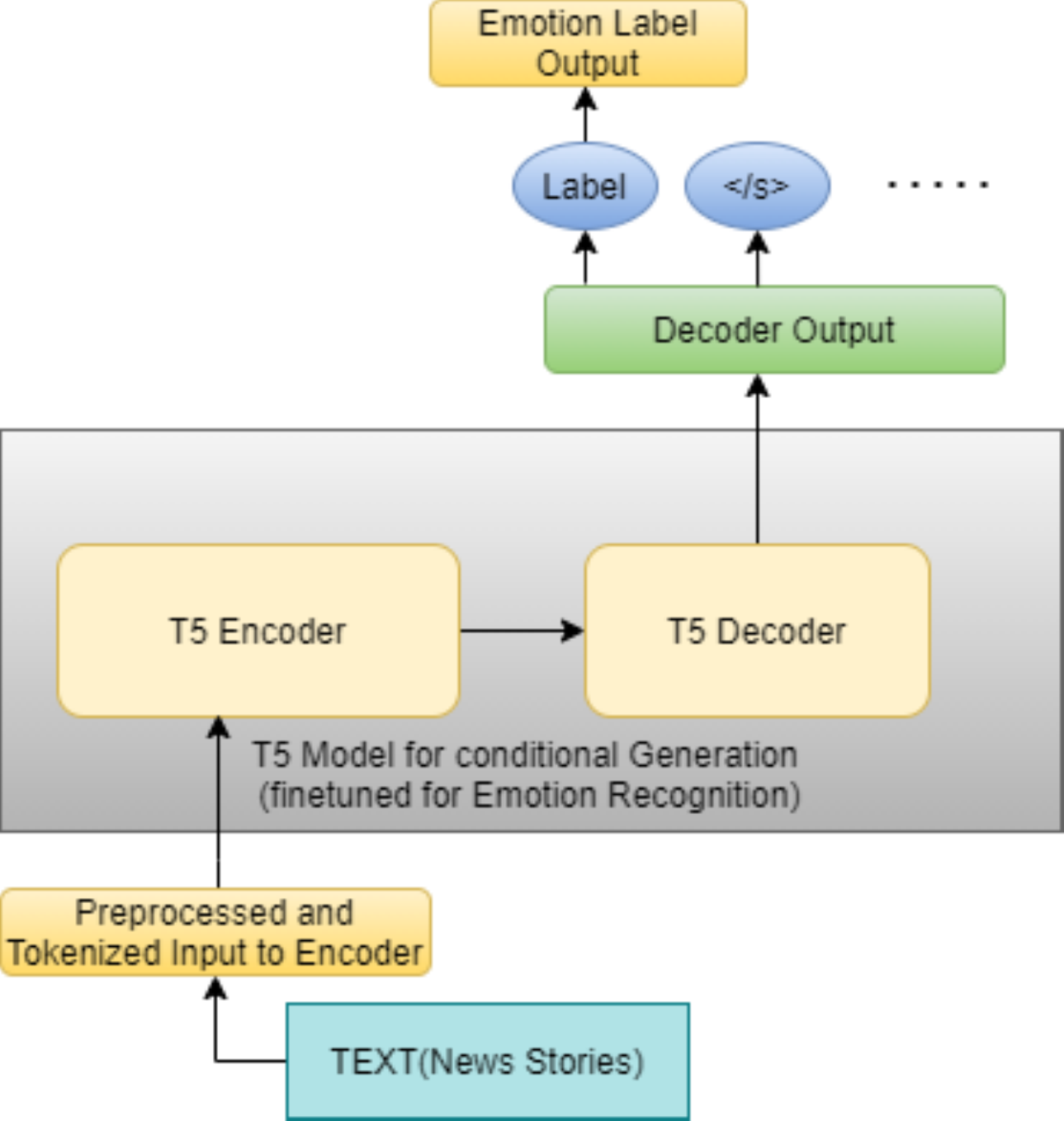}
    \caption{}
    \label{fig:b}
\end{figure}
Our proposed approach for Emotion Prediction is shown in Figure \ref{fig:b}. The approach is primarily based on T5 Model \cite{DBLP:journals/corr/abs-1910-10683} for conditional generation of emotion labels. Hence before feeding into the network, the emotion prediction task is cast as feeding the essay text as input and training it to generate target emotion labels as text. This allows for the use of the same model, loss function, and hyper-parameters for the task of emotion prediction as is done in other Text Generation tasks. More formally, the modeling of the task can be described as:
\[
    p(x|c) = \prod_{i = 1}^{2} p(x_{i}|x_{<i},c) 
\]
where \emph{c} is the encoder input obtained after text tokenization and \emph{x} is the target decoder output which is of length 2, with \emph{x\textsubscript{1}} and \emph{x\textsubscript{2}} as as \emph{label\_token}  and \emph{eos\_token} respectively.

Here, the transformer network parameters $\theta$ are trained with negative log-likelihood over a dataset \emph{D} = \{$(c^{1},x^{1}), (c^{2},x^{2}), ..., (c^{D},x^{D})$\}:

\[
    \mathcal{L}(\mathcal{D}) = - \sum_{k=1}^{|\mathcal{D}|}logp_{\theta}(x_{i}^{k}|x_{<i}^{k},c^{k})
\]

\noindent\textbf{Task Specific Incremental Finetuning}: Along with the architecture proposed, instead of the T5 base model, we propose to use the T5 model finetuned on emotion recognition dataset \cite{saravia-etal-2018-carer} for Emotion Recognition downstream task \cite{T5_finetuned}. This is done to leverage the task-specific knowledge beyond the available dataset of 1860 samples and to analyze the effect of task specific-incremental finetuning. \\

\section{Experiments}

The experiments were conducted using Pytorch \cite{Pytorch} and Hugging Face's transformers \cite{Huggingface}. The experiments were performed on the Google Colab Laboratory tool, which provides a Tesla T4 GPU and 16 GB RAM. Adam's Optimizer \cite{Adam} with a learning rate of 2e-5 was used for optimization. The training batch provided in the WASSA-2021 Shared task consisted of 1860 examples. An 80-20 split on the training set was performed for the train-valid split. The development set consisted of 210 examples was used as a test set. As suggested in the WASSA-2021 Shared Task, the average value of the Pearson correlation score of distress and empathy was used for evaluation in Track 1 and Macro F1 score was used in Track 2.The code and trained models are available at https url\footnote{\url{https://github.com/yashbutala/WASSA}}

\paragraph{Models used in Experiment for Track 1:} For Empathy and Distress Prediction from the finally obtained parameter-rich concatenated vector, we experimented on a variety of Supervised Machine-Learning architectures along with our final system. Brief details on these architectures are given below:
\begin{itemize}
    \item\textbf{SVR} is a Support vector machine that supports linear and non-linear regression. It tries to fit as many instances as possible between the lines while limiting the margin violations. 
    \item\textbf{Ada-Boost} is a meta-estimator that begins by fitting a regressor on the original dataset and then fits additional copies of the regressor on the same dataset but where the weights of instances are adjusted according to the error of the current prediction.
    \item\textbf{XG-Boost} is an applied machine learning algorithm, decision-tree-based ensemble that uses a framework for gradient boosting. It is an implementation of gradient boosted decision trees designed for speed and performance.
    \item\textbf{MLP} is a class of feedforward artificial neural network (ANN). MLP utilizes a supervised learning technique called backpropagation for training.\footnote{Our final approach used for submission in WASSA-2021 Shared Task Track 1} Used the architecture described in the Section \ref{EMP} for final prediction.
    
\end{itemize}

\paragraph{Models used in Experiment for Track 2:} For Emotion Classification, we experimented on a variety of architectures along with our final system. Brief details on these architectures are given below:
\begin{itemize}
    \item\textbf{BERT-base}\cite{DBLP:journals/corr/abs-1810-04805} and \textbf{ALBERT-base-v2}\cite{DBLP:journals/corr/abs-1909-11942} are the text classification models, in which pooled output, (i.e. output from the first token or \emph{[CLS]} token of last layer of the model) is used as the contextualized embeddings for the text which was then fed into a single feed-forward linear layer for Binary Classification\cite{kamal2021hostility}, trained with Binary Cross-Entropy Loss. 
    \item\textbf{T5-base}\cite{DBLP:journals/corr/abs-1910-10683} and \textbf{T5-Finetuned}\cite{T5_finetuned}\footnote{Our final approach used for submission in WASSA-2021 Shared Task Track 2} models use the architecture described in the Section \ref{EMO} for conditional generation, the only difference being the pre-trained versions of T5 model used. 
    \item\textbf{Pegasus-xsum}\cite{DBLP:journals/corr/abs-1912-08777} model also uses the same approach as used in T5 model for conditional generation, and performs better in several downstream tasks. 
\end{itemize}
Links to pre-trained models used for experiment can be found in the Supplemental Section \ref{sec:supplemental}.

\section{Result}
This section discusses the results from different approaches and architectures used in our experiments. While the train and development dataset were already available, the gold standard annotations of the test are withheld in the Shared Task of WASSA-2021. The experiments were performed considering Development set as our test dataset. Performance of only the final submissions are provided on the held-out test dataset.

\label{Result}

\subsection{Results for Track 1}
Table \ref{Table:Result-1} shows the model performance discussed in the development dataset section (used as validation set) for Track 1. Our final approach, which is described in Section \ref{EMP} outperforms other approaches by an appreciable margin on development set but it failed on empathy prediction on the latest test set submission in the Codalab. It was due to an erroneous submission from us. Though during the post-evaluation phase, the model performed better on test set than the \cite{buechel-etal-2018-modeling}. Table \ref{Table:Result-2} shows the performance of our final submission for WASSA-2021 on the held-out test dataset

\begin{table}
\centering
\begin{tabular}{|c|c|c|c|}
 \hline
 \textbf{Predictor}  & \textbf{Distress}  & \textbf{Empathy} & \textbf{Average}\\ 
 \hline
  SVR &  0.400 & 0.406 & 0.403\\
  XG-Boost & 0.431 & 0.394  & 0.413 \\
  Ada-Boost & 0.418 & 0.374 & 0.396\\
  \textbf{MLP*} & \textbf{0.462} & \textbf{0.473} & \textbf{0.468} \\
  \hline
\end{tabular}
\caption{Performance of our system on the development datatset of Track 1. Our final submission approach for Track 1 is marked with *.}
\label{Table:Result-1}
\end{table}

\begin{table}
\centering
\begin{tabular}{|c|c|c|c|}
 \hline
 \textbf{Predictor}  & \textbf{Distress}  & \textbf{Empathy} & \text{Average}\\
 \hline
  MLP & \textbf{0.476} & \textbf{0.358} & \textbf{0.417} \\
 \hline
\end{tabular}
\caption{Performance of our final submission on the held-out Test datatset of Track 1. Our final submission secured \textbf{4\textsuperscript{th}} rank on the Shared Task Leaderboard }
\label{Table:Result-2}
\end{table}

\subsection{Results for Track 2}
Table \ref{Table:Result-Track-2} shows the performance of the models discussed in section on the Development dataset (used as validation set) for Track 2. While its clear that our final approach described in Section \ref{EMO} outperforms other approaches by an appreciable margin, there are two other important aspects to note from the table. First, the performance of Conditional Generation models, i.e. Pegasus-xsum and T5 models over Pre-trained Contextual Embedding based Classification model used for BERT and ALBERT models. Second, the improvement obtained by using already finetuned model\footnote{\url{https://huggingface.co/mrm8488/t5-base-finetuned-emotion}} over T5-base model. As can be seen in Table \ref{Table:Result-Track-2}, our approach of generation of emotion labels performs way better than Contextual Embeddings-based classification. Furthermore, the improvement obtained by using task-specific finetuned model over the base T5 model suggests that the model is able to exploit the benefit of task specific incremental finetuning and was able to extrapolate the knowledge features learned from previous finetuning on Emotion recognition onto the newly finetuned model. Table \ref{Table:Result-Track-2-2} shows the performance of our final submission for WASSA-2021 on the held-out test dataset.
\begin{table}
\centering
\begin{tabular}{|c|c|}
 \hline
 \textbf{Model} & \textbf{Macro-F1 score}\\
 \hline
  BERT-base &  0.38\\
  ALBERT-base-v2 & 0.4739\\
  Pegasus-xsum &  0.502\\
  T5-base &  0.5259\\
  T5-Finetuned* & \textbf{0.572}\\
  \hline
\end{tabular}
\caption{Macro F1 score for Emotion Prediction on Development set. Model descriptions are provided in Section \ref{Result}. Our final submission approach for Track 2 is marked with *.}
\label{Table:Result-Track-2}
\end{table}

\begin{table}
\centering
\begin{tabular}{|c|c|}
 \hline
 \textbf{Metric} & \textbf{Result}\\
 \hline
  Macro F1 Score &  0.502\\
  Micro F1 Score & 0.594\\
  Accuracy &  0.594\\
  Macro Precision &  0.550\\
  Micro Precision &  0.594\\
  Macro Recall &  0.483\\
  Micro Recall & 0.594\\
  \hline
\end{tabular}
\caption{Performance of our system on the held-out test datatset, for track 2. Our final submission secured \textbf{2\textsuperscript{nd}} rank on the Shared Task Leaderboard }
\label{Table:Result-Track-2-2}
\end{table}

\section{Conclusion and Future Work}
In this paper, we presented an approach for predicting empathic concern and personal distress by fine-tuning a pre-trained language model using parameter sharing. 
As empathy and distress are correlated, we observe that parameter sharing improves the performance on this task. 
We amalgamated the sentence embeddings and other additional data, which further used the regression model for prediction. 
The ablation studies show that on the validation set, the MLP works best. 

Also, opposite to most text classification approaches, which use embeddings from the final layer of pre-trained language models, we illustrated the efficiency of formulating finetuning of language models as a label generation task for emotion prediction. Our ablation studies demonstrated usefulness of using task-specific incremental finetuning.

In the future, we plan to extend our multi-tasking based model to incorporate soft parameter sharing. Inclusion of personality related traits provided in the dataset used in this paper, while predicting emotion labels might also be a promising direction to work on.

\newpage

\bibliography{eacl2021}
\bibliographystyle{acl_natbib}

\appendix

\section{Supplemental Material}
\label{sec:supplemental}
Links to the huggingface models used in the experiment:
\begin{itemize}
    \item Bert-base: \url{https://huggingface.co/bert-base-uncased}
    \item Albert-base-v2: \url{https://huggingface.co/albert-base-v2}
    \item Pegasus-xsum: \url{https://huggingface.co/google/pegasus-xsum}
    \item T5-base: \url{https://huggingface.co/t5-base}
    \item T5-Finetuned: \url{https://huggingface.co/mrm8488/t5-base-finetuned-emotion}
\end{itemize}

\end{document}